\def\year{2021}\relax
\documentclass[letterpaper]{article} 
\usepackage{aaai21}  
\usepackage{times}  
\usepackage{helvet} 
\usepackage{courier}  
\usepackage[hyphens]{url}  
\usepackage{graphicx} 
\usepackage{natbib}  
\usepackage{caption} 
\usepackage{multirow}
\usepackage[table,xcdraw]{xcolor}
\usepackage{booktabs}
\usepackage[ruled, linesnumbered]{algorithm2e}
\usepackage{bm}

\usepackage{times}
\usepackage{latexsym}

\usepackage{amsmath}
\usepackage{amssymb} 
\usepackage{booktabs}
\usepackage{amsfonts}
\usepackage{bm}
\usepackage{multirow}
\usepackage{soul}
\usepackage{url}
\usepackage[hidelinks]{hyperref}
\usepackage[utf8]{inputenc}
\usepackage{caption}

\title{Spatial-Temporal Fusion Graph Neural Networks for Traffic Flow Forecasting}
\author{
	Mengzhang Li, Zhanxing Zhu\\
}
\affiliations{
    Peking University, Beijing, China

    \{mcmong, zhanxing.zhu\}@pku.edu.cn

}

\begin{document}

\maketitle

\begin{abstract}
Spatial-temporal data forecasting of traffic flow is a challenging task because of complicated spatial dependencies and dynamical trends of temporal pattern between different roads. Existing frameworks typically utilize given spatial adjacency graph and sophisticated mechanisms for modeling spatial and temporal correlations. However, limited representations of given spatial graph structure with incomplete adjacent connections may restrict effective spatial-temporal dependencies learning of those models. Furthermore, existing methods are out at elbows when solving complicated spatial-temporal data: they usually utilize separate modules for spatial and temporal correlations, or they only use independent components capturing localized or global heterogeneous dependencies. To overcome those limitations, our paper proposes a novel Spatial-Temporal Fusion Graph Neural Networks (STFGNN) for traffic flow forecasting. First, a data-driven method of generating “temporal graph” is proposed to compensate several existing correlations that spatial graph may not reflect. SFTGNN could effectively learn hidden spatial-temporal dependencies by a novel fusion operation of various spatial and temporal graphs, treated for different time periods in parallel. Meanwhile, by integrating this fusion graph module and a novel gated convolution module into a unified layer, SFTGNN could handle long sequences by learning more spatial-temporal dependencies with layers stacked. Experimental results on several public traffic datasets demonstrate that our method achieves state-of-the-art performance consistently than other baselines\footnote{Code available at: https://github.com/MengzhangLI/STFGNN}.
\end{abstract}

\section{Introduction}
Forecasting task of spatial-temporal data especially traffic data has been widely studied recently because (1) traffic forecasting is one of the most important part of Intelligent Transportation System (ITS) which has great effect on daily life; (2) its data structures is also representative in reality: other location-based data such as wind energy stations, air monitoring stations and cell towers can all be formulated as this spatial-temporal data structure. 

Recently, graph modeling on spatial-temporal data has been in the spotlight with the development of graph neural networks. Many works have achieved impressive performance on prediction accuracy. Although significant improvements have been made in incorporating graph structure into spatial-temporal data forecasting model, these models still face several shortcomings. 

\begin{figure}
	\centering
	\includegraphics[height=3in]{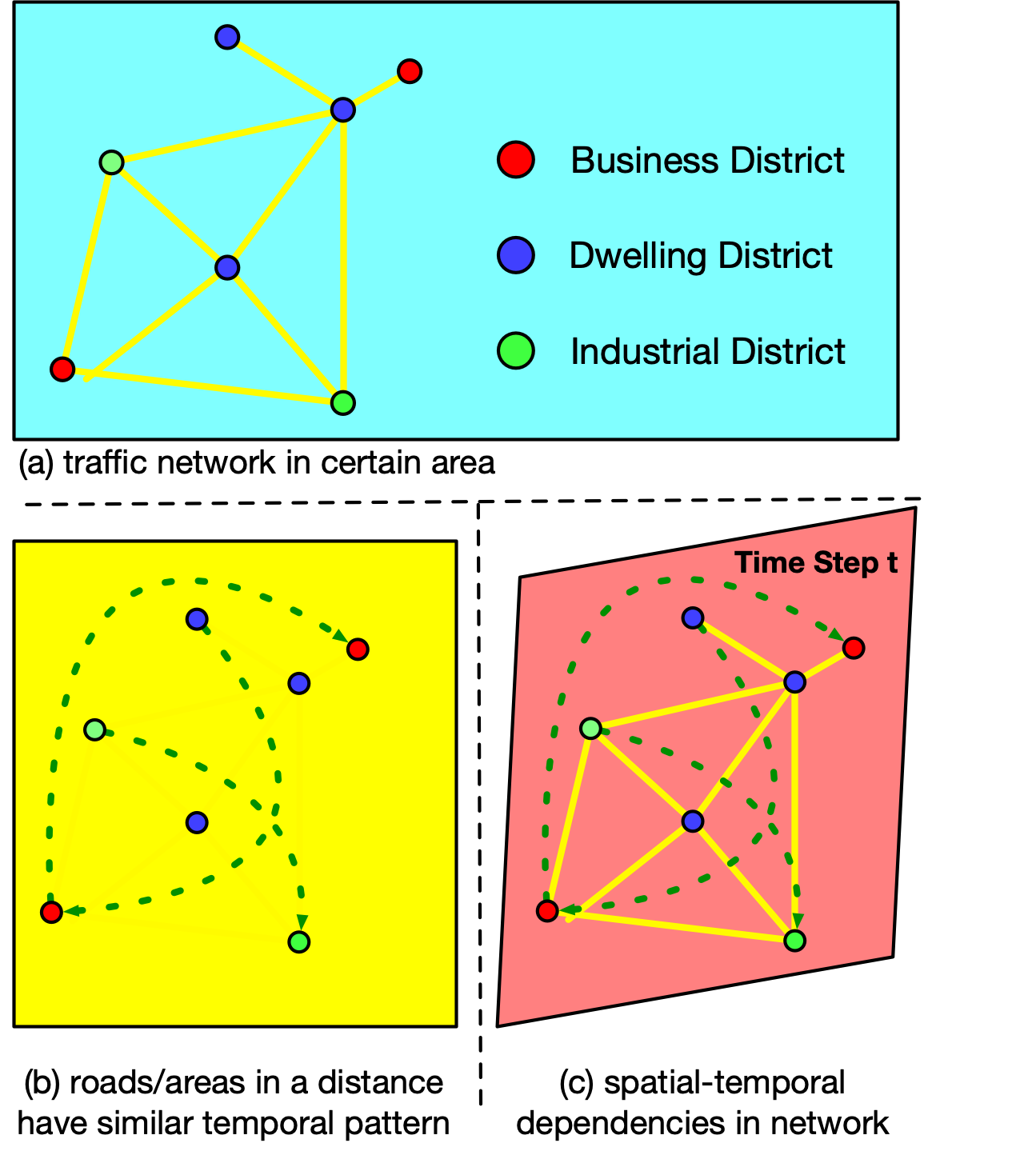}
	\caption{Example of spatial-temporal dependencies in a network. Yellow lines indicate the spatial adjacency in reality. Districts play the same role in traffic network are likely to have similar temporal pattern, which are represented by green dash lines.}
	\label{fig:sp_relation_example}
\end{figure}

The first limitation is being lack of an informative graph construction. Taking  Figure \ref{fig:sp_relation_example} for example, those distant nodes may have certain correlations, i.e., they would share similar "temporal pattern". For instance, during rush hours, most roads near the office buildings (from business districts) will encounter traffic jams in the same period. However, most existing models only utilize given spatial adjacency matrix for graph modeling, and \emph{ignore the temporal similarity between nodes when modeling the adjacency matrix}. Some works already made several attempts to improve representation of graph. Mask matrix \cite{song2020spatial} and self-adaptive matrix \cite{wu2019graph} are introduced to adjust existed spatial adjacency matrix, but these learnable matrices are both lack of correlations representation ability for complicated spatial-temporal dependencies in graph. Temporal self-attention module  \cite{xu2020spatial,wang2020traffic} of transformers can also extract dynamic spatial-temporal correlations, and predetermined spatial graph may not reflect it. However, it may face overfitting of spatial-temporal dependencies learning due to dynamical change and noisy information in reality data especially in long-range prediction tasks, where autoregressive models can hardly avoid error accumulation.


Besides, current studies of spatial-temporal data forecasting are \emph{ineffective to capture dependencies between local and global correlations}. RNN/LSTM-based models \cite{li2017diffusion,zhang2018gaan} are time-consuming and may suffer gradient vanishing or explosion when capturing long-range sequences. Sequential procedure of transformers~\cite{park2019stgrat,wang2020traffic,xu2020spatial} may still be time-consuming in inference. CNN-based methods need to stack layers for capturing global correlations of long sequences. STGCN~\cite{yu2017spatio} and GraphWaveNet~\cite{wu2019graph} may lose local information if dilation rate increases. STSGCN~\cite{song2020spatial} proposes a novel localized spatial-temporal subgraph that synchronously capture local correlations, which is only designed locally and ignoring global information. When missing data happens, situation is more severe where it would only learn local noise.

To capture both local and global complicated spatial-temporal dependencies, we present a novel CNN-based framework called Spatial-Temporal Fusion Graph Neural Network (STFGNN). Motivated by dynamic time warping~\cite{berndt1994using}, we propose a novel data-driven method for graph construction: the temporal graph learned based on similarities between time series. Then several graphs could be integrated as a spatial-temporal fusion graph to  obtain hidden spatial-temporal dependencies. Moreover, to break the local and global correlation trade-off, gated dilated convolution module is introduced, whose larger dilation rate could capture long-range dependencies. The main contributions of this work are as follows.

\begin{itemize}
\item We construct a novel graph by a data-driven method, which preserve hidden spatial-temporal dependencies. This data-driven adjacency matrix is able to extract correlations that given spatial graph may not present. Then, we propose a novel spatial-temporal fusion graph module to capture spatial-temporal dependencies synchronously. 

\item We propose an effective framework to capture local and global correlations simultaneously, by assembling a Gated dilated CNN module with spatial-temporal fusion graph module in parallel. Long-range spatial-temporal dependencies could also be extracted with layers stacked.

\item To make thorough comparisons and test performance in complicated cases, extensive experiments are conducted on four real-world datasets used in previous works, respectively. The results show our model consistently outperforms baselines, which strongly proves our proposed model could handle complicated traffic situations in reality with different traffic characteristics, road numbers and missing value ratios.
\end{itemize}


\begin{figure*}[!htb]
	\centering
	\includegraphics[width=4.5in]{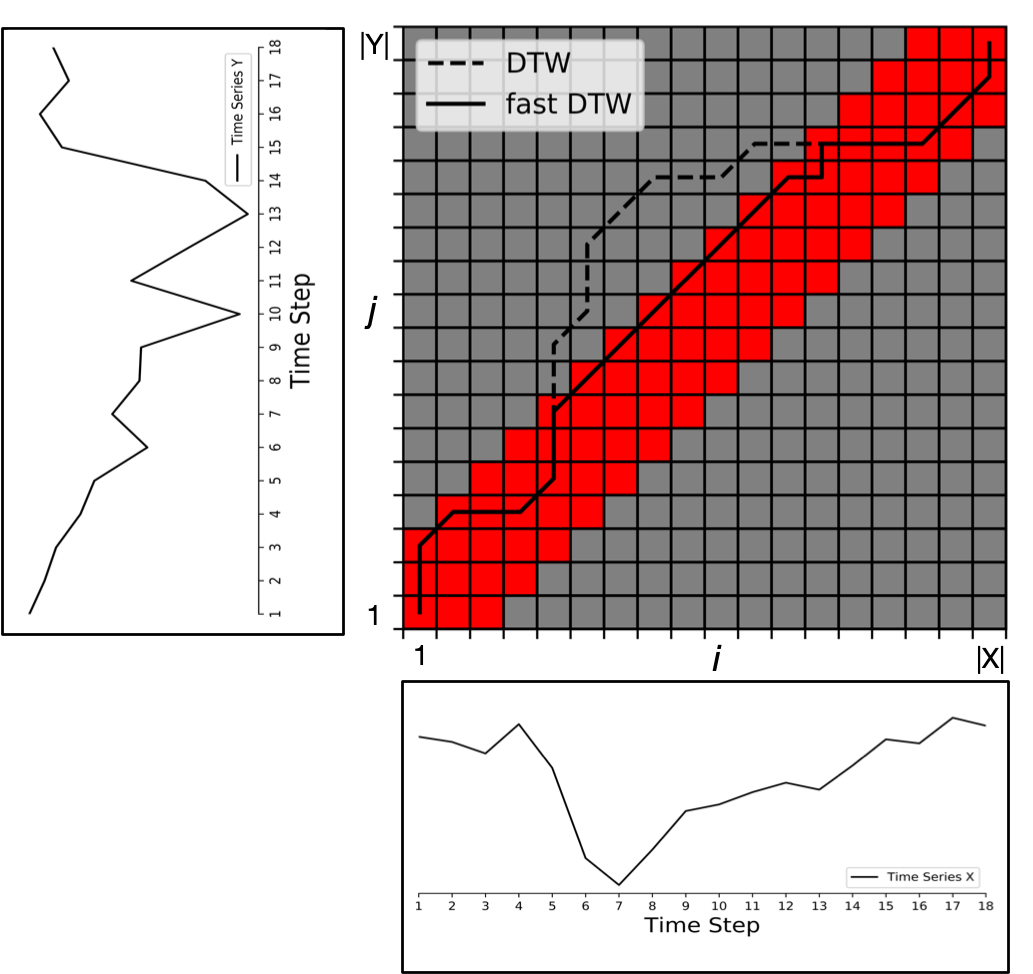}
	\caption{Two time series and their warping path calculated by DTW and fast-DTW algorithm. The red zone is searching zone of fast-DTW defined by "Searching Length" $T$.}
	\label{fig:dtw}
\end{figure*}

\section{Related Works}
\subsection{Graph Convolution Network}
Graph convolution networks are widely applied in many graph-based tasks such as classification \cite{kipf2016semi} and clustering \cite{chiang2019cluster}, which has two types. One is extending convolutions to graphs in spectral domain by finding the corresponding Fourier basis \cite{bruna2013spectral}. GCN \cite{kipf2016semi} is representative work and constructs typical baselines in many tasks. The other is generalizing spatial neighbours by typical convolution. GAT \cite{velivckovic2017graph} which introduces attention mechanism into graph filed, and GraphSAGE \cite{hamilton2017inductive} which generates node embeddings by sampling and aggregating features locally are all typical works.

\subsection{Spatial-Temporal Forecasting}
Spatial-temporal prediction plays an important role in many application areas. To incorporate spatial dependencies more effectively, recent works introduce graph convolutional network (GCN) to learn the traffic networks. DCRNN \cite{li2017diffusion} utilizes the bi-directional random walks on the traffic graph to model spatial information and captures temporal dynamics by gated recurrent units (GRU). Transformer models such as \cite{wang2020traffic,park2019stgrat} utilize spatial and temporal attention modules in transformer for spatial-temporal modeling. They would be more effective when training than LSTM but still make predictions step by step due to their autoregressvie structures. STGCN \cite{yu2017spatio} and GraphWaveNet \cite{wu2019graph} employed graph convolution on spatial domain and 1-D convolution along time axis. They process graph information and time series separately. STSGCN \cite{song2020spatial} make attempts to incorporate spatial and temporal blocks altogether by localized spatial-temporal synchronous graph convolution module regardless of global mutual effect. 

\begin{algorithm}
	\small{
	\caption{\label{alg:tgg} Temporal Graph Generation}
	\KwIn{ N time series from $\mathcal{V} ( \vert \mathcal{V} \vert = N)$}
	$W$ Initialization, reset to zero matrix
	TDL: Temporal Distance Calculation defined in Alg \ref{alg:tdl}\\
	\For{$i = 1, 2, \cdots, N$}{
		\For{$j = 1, 2, \cdots, N$}{
			$dist_{i, j}$ = TDL($V_i, V_j$) (Alg. \ref{alg:tdl})
		}
		Sort smallest $k (k \leq N)$ element and their index $\mathbf{j} = \{ j_1, j_2, \cdots, j_k \}$ s.t. $dist_{i, j_1} \leq dist_{i, j_2} \leq dist_{i, j_k}$
		\lIf{$\tilde{j} \in \mathbf{j}$}{$W_{i, \tilde{j}} = W_{\tilde{j}, i} = 1$}
	}
	\Return Weighted Matrix $W$ of Temporal Graph $\mathcal{G}$.
	}
\end{algorithm}

\subsection{Similarity of Temporal Sequences}
The methods for measuring the similarity between time series can be divided into three categories: (1) timestep-based, such as Euclidean distance reflecting point-wise temporal similarity; (2) shape-based, such as Dynamic Time Warping \cite{berndt1994using} according to the trend appearance; (3) change-based, such as Gaussian Mixture Model(GMM) \cite{povinelli2004time} which reflects similarity of data generation process. 

Dynamic Time Warping is a typical algorithm to measure similarity of time series. Given two time series $X = (x_1, x_2, \cdots, x_n)$ and $Y = (y_1, y_2, \cdots, y_m)$, series distance matrix \bm{$M_{n\times m}$} could be introduced whose entry is $M_{i, j} = \vert x_i - y_j \vert$. Then cost matrix \bm{$M_c$} could be defined:
\begin{small}
	\begin{equation}
	M_c(i,j) = M_{i,j} + \min(M_c(i, j-1), M_c(i-1, j), M_c(i,j))\label{mc}
	\end{equation}
\end{small}

After several iterations of $i$ and $j$, $dist(X, Y) = M_c(n, m)^{\frac{1}{2}}$ is the final distance between $X$ and $Y$ with the best alignment which can represent the similarity between two time series.

From Eq. (\ref{mc}) we can tell that Dynamic Time Warping is an algorithm based on dynamic programming and its core is solving the warping curve, i.e., matchup of series points $x_i$ and $y_j$. In other words the "warping path" \bm{$\Omega$}
\begin{align}
\Omega = (\omega_1, \omega_2, \cdots, \omega_\lambda), \quad \max(n, m) \leq \lambda \leq n+m\notag
\end{align}
is generated through iterations of Eq. (\ref{mc}). Its element $\omega_\lambda = (i, j)$ means matchup of $x_i$ and $y_j$.

\section{Preliminaries}
We can represent the road network as a graph $\mathcal{G} = (V, E, {A}_{SG})$, where $V$ is a finite set of nodes $\vert V\vert = N$\footnote{In this paper, $N$ represents number of traffic roads/nodes, $n$ represents given length of certain time series. They are totally different.}, corresponding to the observation of $N$ sensors or roads; $E$ is a set of edges and ${A}_{SG} \in \mathbb{R}^{N \times N}$ is a spatial adjacency matrix representing the nodes proximity or distance. Denote the observed graph signal ${X}_{\mathcal{G}}^{(t)} \in \mathbb{R}^{N \times d}$ means it represent the observation of spatial graph information $\mathcal{G}$ at time step $t$, whose element is observed $d$ traffic features(e.g., the speed, volume) of each sensor. The aim of traffic forecasting is learning a function $f$ from previous $T$ speed observations to predict next $T^{'}$ traffic speed from $N$ correlated sensors on the road network.
\begin{equation}
[\mathbf{X}_{\mathcal{G}}^{(t-T+1)}, \cdots, \mathbf{X}_{\mathcal{G}}^{t}] \xrightarrow[]{f} [ \mathbf{X}^{t+1}_{\mathcal{G}}, \cdots, \mathbf{X}^{t+T^{'}}_{\mathcal{G}}]\label{d1}
\end{equation}

\begin{algorithm}
	\small{
		\caption{\label{alg:tdl} Temporal Distance Calculation (TDL)}
		\KwIn{$X = (x_1, \cdots, x_n) \in \mathbb{R}^{n \times d}$, $Y = (y_1, \cdots, y_m) \in \mathbb{R}^{m \times d}$, Searching Length $T$}		
		\For{$i = 1, 2, \cdots, n$}{
			\For{$j = \max(0, i-T), \cdots, \min(m, i + T +1)$}{
				$M_{i, j} = \vert X_i - Y_j \vert$\;
				\lIf{$i = 0, j = 0$}{$M_C(i, j) = M_{i, j}^2$}
				\lElseIf{$i = 0$}{$M_C(i, j) = M_{i, j}^2 + M_{i, j-1}$}
				\lElseIf{$j = 0$}{$M_C(i, j) = M_{i, j}^2 + M_{i-1, j}$}
				\lElseIf{$j = i - T$}{$M_C(i, j) = M_{i, j}^2 + \min(M_{i-1, j-1}, M_{i-1, j} )$}
				\lElseIf{$j = i + T$}{$M_C(i, j) = M_{i, j}^2 + \min(M_{i-1, j-1}, M_{i, j-1} )$}
				\lElse{$M_C(i, j) = M_{i, j}^2 + \min(M_{i-1, j-1}, M_{i, j-1}, M_{i-1, j})$}
			}
		}
		\Return $dist(X, Y) = M_C(n, m)^{\frac{1}{2}}$
	}
\end{algorithm}

\section{Spatial-Temporal Fusion Graph Neural Networks}
We present the framework of Spatial-Temporal Fusion Graph Neural Network in Figure \ref{fig:stfgnn}. It consists of (1) an input layer, (2) stacked Spatial-Temporal Fusion Graph Neural Layers and (3) an output layer. The input and output layer are one and two Fully-Connected Layer followed by activation layer such as "ReLU" \cite{nair2010rectified} respectively. Every Spatial-Temporal Fusion Graph Layer is constructed by several Spatial-Temporal Fusion Graph Neural Modules (STFGN Modules) in parallel and a Gated CNN Module which includes two parallel 1D dilated convolution blocks. 

\subsection{Spatial-Temporal Fusion Graph Construction}
The aim of generating temporal graph is to achieve certain graph structure with more accurate dependency and genuine relation than spatial graph. Then, incorporating temporal graph into a novel spatial-temporal fusion graph, which could \textbf{make deep learning model lightweight} because this fusion graph already has correlation information of each node with its (1) spatial neighbours, (2) nodes with similar temporal pattern, and (3) own previous or later situation along time axis.

However, generating temporal graph based on similarity of time series by DTW is not easy, it is a typical dynamic programming algorithm with computational complexity $\mathcal{O}(n^2)$. Thus it might be unacceptable for many applications because time series of real world is usually very long. To reduce complexity of DTW, we restrict its "Search Length" $T$. The searching space of warping path is circumscribed by:
\begin{equation}
\omega_k = (i, j), \quad \vert i - j \vert \leq T
\end{equation}
Consequently, the computational complexity of DTW is reduced from $\mathcal{O}(n^2)$ to $\mathcal{O}(T n)$ which made its application on large scale spatial-tempral data possible. We name it "fast-DTW".

As shown in Figure \ref{fig:dtw}, given two roads' time series whose length is $\vert X\vert$ and $\vert Y \vert$, repectively. The distance of those two time series $M_c(\vert X \vert, \vert Y \vert)$ could be calculated by Eq. (\ref{mc}). The warping path of fast-DTW is restrcited near the diagonal (red zone in Figure \ref{fig:dtw}), consequently the cost of calculating match $i$ and $j$ of the element $\omega_\lambda = (i, j)$ of warping path \bm{$\Omega$} is not as expensive as DTW algorithm.

The set of $\alpha$ which determines how many smallest numbers are treated in Alg. \ref{alg:tgg} is tricky and we would analysis it in the section of experiments. Empirically, we keep the sparisity of Temporal Graph $A_{TG}$ almost the same as Spatial Graph $A_{SG}$.

Figure \ref{fig:stfgnn}(b) is the example of Spatial-Temporal Fusion Graph. It consists of three kinds $N \times N$ matrix: Spatial Graph $A_{SG}$ which is given by dataset, Temporal Graph $A_{TG}$ generated by Alg. \ref{alg:tgg}, and Temporal Connectivity graph $A_{TC}$ whose element is nonzero iif previous and next time steps is the same node. Given Spatial-Temporal Fusion Graph $A_{STFG} \in \mathbb{R}^{3N \times 3N}$, and taken $A_{TG}$ within a red circle in Figure \ref{fig:stfgnn}(b) for instance. It denotes the connection between same node from time step: 2 to 3 (current time step $t$ = 2). For each node $l \in \{1, 2, \cdots, N\}$, $i = (t+1)*N + l = 3N + l$ and $j = t * N + l = 2N + l$, then $A_{STFG (i, j)} = 1$. To sum up, Temporal Connectivity graph denotes connection of the same node at proximate time steps.

Finally, Spatial-Temporal Fusion Graph $A_{STFG} \in \mathbb{R}^{KN \times KN}$ is generated. Altogether with the sliced input data of each STFGN Module: 
\begin{align}
h^0 = [{X}_{\mathcal{G}}^{(t)}, \cdots, {X}_{\mathcal{G}}^{(t + K)} ]\in \mathbb{R}^{K \times N \times d \times C}
\end{align}
It is sliced iteratively from total input data: 
\begin{align}
X = [{X}_{\mathcal{G}}^{(t)}, \cdots, {X}_{\mathcal{G}}^{(t + T)} ]\in \mathbb{R}^{T \times N \times d \times C}
\end{align}

${X}_{\mathcal{G}}^{(t)}$ is high-dimension feature of original data $\mathbf{X}_{\mathcal{G}}^{(t)}$. $C$ is the number of input feature channel from STFGN module, which is also the number of output feature channel from input layer.

\subsection{Spatial-Temporal Fusion Graph Neural Module}
\begin{figure*}[!htb]
	\centering
	\includegraphics[width=6in]{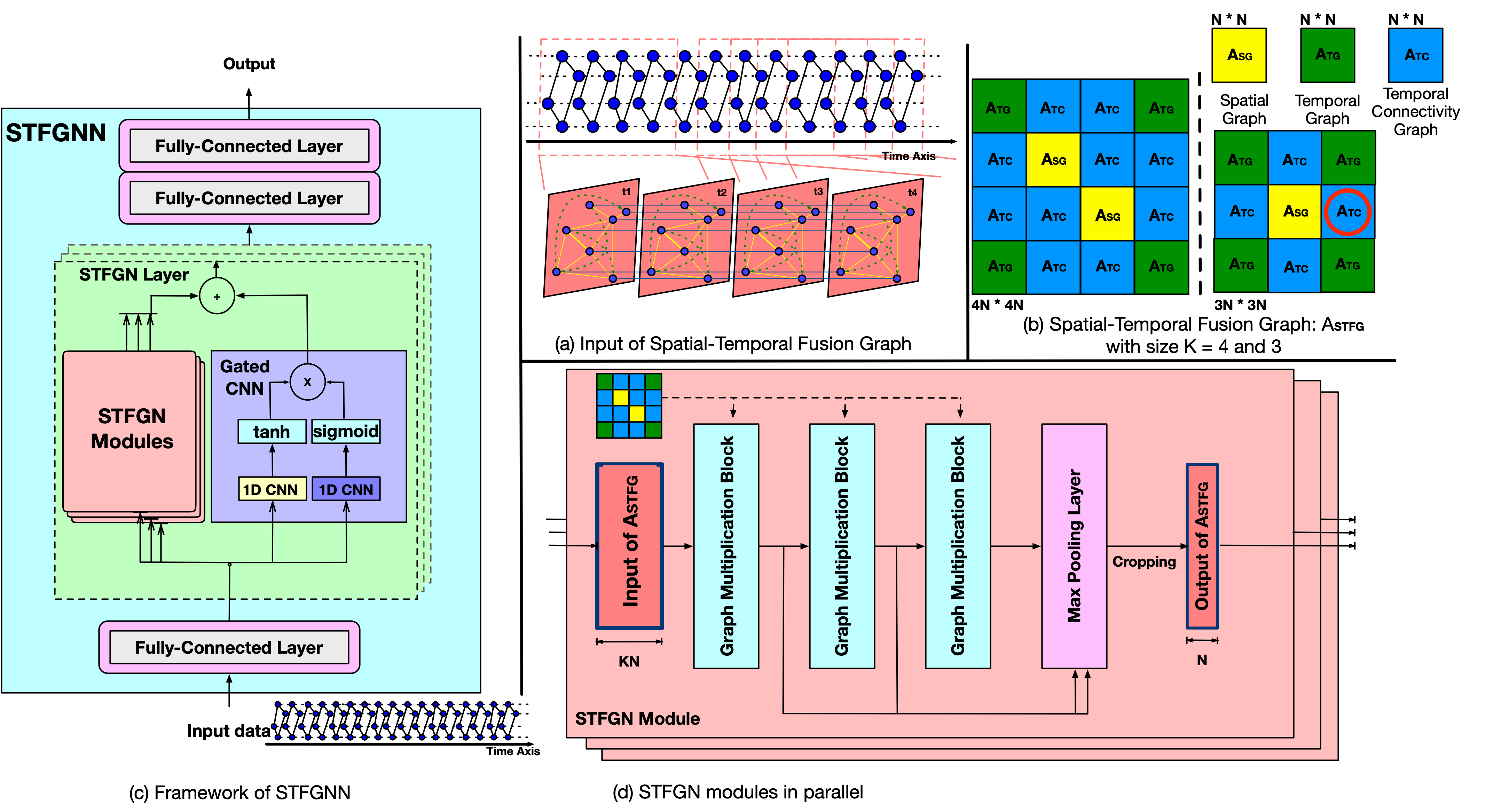}
	\caption{Detailed framework of STFGNN. (a) is the example of input of Spatial-Temporal Fusion Graph, which would be generated iteratively along the time axis. (b) is the example of Spatial-Temporal Fusion Graph, whose size K is 4 and 3, respectively. It consists of three kinds of adjacency matrix $\in N \times N$: spatial graph $A_{SG}$, temporal graph $A_{TG}$ and temporal connectivity graph $A_{TC}$. The $A_{TC}$ within a red circle would be taken for instance in the body. (c) is overall structure of STFGNN, its Gated CNN module and STFGNN modules are in parallel. (d) is detailed architecture of the Spatial-Temporal Fusion Graph Modules, each module will be independently trained for input iteratively generated from (a) in parallel as well.}
	\label{fig:stfgnn}
\end{figure*}

In Spatial-Temporal Fusion Graph Neural Module (STFGN Module), the lightweight deep learning model could extract hidden spatial-temporal dependencies by several simple operations such as matrix multiplication with Spatial-Temporal Fusion Graph $A_{STFG}$, residual connections and max pooling.

In this paper, regular spectral filter such as Laplacian in graph convolution is replaced with a more simplified and time-saving operation: matrix multiplication. Each node in network could aggregate spatial dependency from $A_{SG}$, temporal pattern correlation from $A_{TG}$ and its own proximate correlation long time axis from $A_{TC}$ by several times matrix multiplication with $A_{STFG}$. 

Gating mechanism in LSTM/RNN is also utilized in graph multiplication block. In STFGN Module, gated linear units is used for generalization in graph multiplication by its nonlinar activation. Graph multiplication module is formulated as below:

\begin{align}
h^{l+1} = (A^* h^{l} W_1 + b_1) \odot \sigma(A^* h^l W_2 + b_2)\label{STFG_GLU}
\end{align}
where $h^{l}$ denotes $l$-th hidden states of certain STFGN module. $A^*$ is shorthand of spatial-temporal fusion graph $A_{STFG} \in \mathbb{R}^{KN \times KN}$, $W_1, W_2 \in \mathbb{R}^{C \times C}$, $b_1, b_2 \in \mathbb{R}^{C}$ are all model parameters of GLU. $\odot$ means Hadamard product and $\sigma$ means sigmoid function.

By stacking L graph multiplication blocks, more complicated and non-local spatial-dependencies could be aggregated. Intuitively, the residual connections \cite{he2016deep} would also be introduced for each block. Max Pooling would be operated on the concatenation of each hidden state $ h^M = MaxPool([ h^1, \cdots, h^L ]) \in \mathbb{R}^{K\times N\times d \times C}$. Finally this concatenation corresponding to the middle time step would be cropped, saving $$h^o = h^M\big [\lfloor \frac{K}{2} \rfloor :\lfloor \frac{K}{2} \rfloor+1, :,:,: \big ] \in \mathbb{R}^{1 \times N \times d \times C}$$ 

Figure \ref{fig:stfgnn}(b) shows this cropped feature has contained complicated heterogeneity. In each matrix multiplication, $A_{SG}$ in the middle of diagonal (corresponding to cropped location of concatenation) transmit information from spatial neighbour. $A_{TC}$ in its horizontal and vertical direction gives each node its own information along time axis. $A_{TG}$ in corner enhance information from nodes with similar temporal pattern.

Input data would be treated by multiple STFGN Modules independently in parallel, which is time-saving and could capture more complicated correlations. Then concatenation of each STFGN module output would be added with Gated CNN output and becomes input of next STFGN layer. Noted that the size of each STFGN module output is $\mathbb{R}^{(T-K+1) \times N \times d\times C}$, i.e., each STFGN layer would cut input from $T$ to $T - K + 1$ in time dimensions. It means STFGN layers could stack up to $\lfloor \frac{T}{K-1}\rfloor - 1$ layers.

\subsection{Gated Convolution Module}
Although $A_{STFG}$ could extract global spatial-temporal dependencies by integration of $A_{TG}$, the correlation it contains is more from nodes in a distant (like the example from Figure \ref{fig:sp_relation_example}). Long-range spatial-temporal dependencies of the node itself is also important, which is very challenging for many CNN-based works \cite{yu2017spatio,wu2019graph,song2020spatial} because inborn structure of CNN can hardly outperform auto-regressive models like transformer\cite{park2019stgrat,wang2020traffic}. Different from previous work like GraphWaveNet and STGCN, dilated convolution with large dilation rate is introduced in this paper. Given the total input data $X \in \mathbb{R}^{T \times N \times d \times C}$, it takes the form:
\begin{align}
Y = \phi(\Theta_1 \ast X + a) \odot \sigma(\Theta_2 \ast X + b) \label{CNN_GLU}
\end{align}

Similar with Eq. (\ref{STFG_GLU}), $\phi(\cdot)$ and $\sigma(\cdot)$ are tanh and sigmoid function, respectively. $\Theta_1$ and $\Theta_2$ are two independent 1D convolution operation with dilation rate = K -1. It could enlarge receptive filed along time axis thus strengthen model performance for extracting sequential dependencies.

Huber loss is chosen as loss function, objective function is shown below:
\begin{align}
L(\mathbf{\hat{X}}_{\mathcal{G}}^{(t+1) : (t+T)}, \Theta) = \frac{\sum\limits_{i=1}^{T} \sum\limits_{j=1}^{N} \sum\limits_{k=1}^{d} h\big ( \mathbf{\hat{X}}_{\mathcal{G}}^{(t+i)}, \mathbf{X}_{\mathcal{G}}^{(t+i)} \big )}{T\times N \times d}
\end{align}

$$ h\big ( \hat{Y}, Y \big ) =\left\{
\begin{aligned}
&\frac{1}{2} (\hat{Y} - Y)^2, & \vert \hat{Y} - Y \vert \leq \delta \\
&\delta \vert \hat{Y} - Y \vert - \frac{1}{2} \delta^2, & \vert \hat{Y} - Y \vert > \delta
\end{aligned}
\right.
$$
$\delta$ is hyperparameter to control sensitivity of squared error loss.

\section{Experiments}
\subsection{Datasets}

\begin{table}[]
	\scalebox{0.9}{
	\begin{tabular}{ccccc}
		\hline
		Datasets  & \#Nodes & \#Edges & \#TimeSteps &\#MissingRatio \\ \hline
		PEMS03    & 358     & 547     & 26208     & 0.672\%       \\
		PEMS04    & 307     & 340     & 16992     & 3.182\%        \\
		PEMS07    & 883     & 866     & 28224     & 0.452\%        \\
		PEMS08    & 170     & 295     & 17856     & 0.696\%        \\ \hline
	\end{tabular}}
	\caption{Dataset description and statistics.}
	\label{tab:data_info}
\end{table}

We verify the performance of STFGNN on four public traffic network datasets. PEMS03, PEMS04, PEMS07, PEMS08 released by\cite{song2020spatial}. Those four datasets are constructed from four districts, respectively in California. All these data is collected from the Caltrans Performance Measurement System (PeMS) and aggregated into 5-minutes windows, which means there are 288 points in the traffic flow for one day. The spatial adjacency networks for each dataset is constructed by actual road network based on distance. Z-score normalization is adopted to standardize the data inputs. The detailed information is shown in Table \ref{tab:data_info}.

\begin{table*}[!htb]
	\centering
	\scalebox{0.88}{
		\begin{tabular}{ccccccccc}
			\hline
			\multirow{2}{*}{\shortstack{Datasets}} &  \multirow{2}{*}{\shortstack{Metric}}                            & \multirow{2}{*}{\shortstack{FC-LSTM}} & \multirow{2}{*}{DCRNN} & \multirow{2}{*}{STGCN} & \multirow{2}{*}{\shortstack{ASTGCN(r)}} & \multirow{2}{*}{\shortstack{Graph WaveNet}} & \multirow{2}{*}{STSGCN} & \multirow{2}{*}{STFGNN} \\
			&                                              &                          &                        &                        &                            &                               &                         &                         \\ \hline
			\multicolumn{1}{c|}{\multirow{3}{*}{\shortstack{PEMS03}}} & \multicolumn{1}{c|}{MAE}                      & 21.33 $\pm$ 0.24                    & 18.18 $\pm$ 0.15                  & 17.49 $\pm$ 0.46                  & 17.69 $\pm$ 1.43                      & \textit{19.85 $\pm$ 0.03}                & 17.48 $\pm$ 0.15                   & \textbf{16.77 $\pm$ 0.09}          \\ 
			\multicolumn{1}{c|}{}                        & \multicolumn{1}{c|}{MAPE(\%)}                & 23.33 $\pm$ 4.23                    & 18.91 $\pm$ 0.82                  & 17.15 $\pm$ 0.45                  & 19.40 $\pm$ 2.24                      & \textit{19.31 $\pm$ 0.49}                & 16.78 $\pm$ 0.20                  & \textbf{16.30$\pm$ 0.09}          \\ 
			\multicolumn{1}{c|}{}                        & \multicolumn{1}{c|}{RMSE}                & 35.11 $\pm$ 0.50                   & 30.31 $\pm$ 0.25                  & 30.12 $\pm$ 0.70                 & 29.66 $\pm$ 1.68                      & \textit{32.94 $\pm$ 0.18}                & 29.21 $\pm$ 0.56                   & \textbf{28.34$\pm$ 0.46}          \\ \hline
			\multicolumn{1}{c|}{\multirow{3}{*}{\shortstack{PEMS04}}} & \multicolumn{1}{c|}{MAE}                & 27.14 $\pm$ 0.20                   & 24.70 $\pm$ 0.22                  & 22.70 $\pm$ 0.64                  & 22.93 $\pm$ 1.29                      & \textit{25.45 $\pm$ 0.03}                & 21.19 $\pm$ 0.10                  & \textbf{19.83$\pm$ 0.06}          \\ 
			\multicolumn{1}{c|}{}                        & \multicolumn{1}{c|}{MAPE(\%)}                & 18.20 $\pm$ 0.40                    & 17.12 $\pm$ 0.37                  & 14.59 $\pm$ 0.21                 & 16.56 $\pm$ 1.36                      & \textit{17.29 $\pm$ 0.24}                & 13.90 $\pm$ 0.05                   & \textbf{13.02$\pm$ 0.05}          \\ 
			\multicolumn{1}{c|}{}                        & \multicolumn{1}{c|}{RMSE}                & 41.59 $\pm$ 0.21                   & 38.12 $\pm$ 0.26                  & 35.55 $\pm$ 0.75                  & 35.22 $\pm$ 1.90                      & \textit{39.70 $\pm$ 0.04}                & 33.65 $\pm$ 0.20                   & \textbf{31.88$\pm$ 0.14}          \\ \hline
			\multicolumn{1}{c|}{\multirow{3}{*}{\shortstack{PEMS07}}} & \multicolumn{1}{c|}{MAE}                & 29.98 $\pm$ 0.42                    & 25.30 $\pm$ 0.52                  & 25.38 $\pm$ 0.49                 & 28.05 $\pm$ 2.34                      & \textit{26.85 $\pm$ 0.05}                & 24.26 $\pm$ 0.14                   & \textbf{22.07$\pm$ 0.11}          \\ 
			\multicolumn{1}{c|}{}                        & \multicolumn{1}{c|}{MAPE(\%)}                & 13.20 $\pm$ 0.53                    & 11.66 $\pm$ 0.33                  & 11.08 $\pm$ 0.18                 & 13.92 $\pm$ 1.65                     & \textit{12.12 $\pm$ 0.41}                & 10.21 $\pm$ 1.65                   & \textbf{9.21$\pm$ 0.07}           \\ 
			\multicolumn{1}{c|}{}                        & \multicolumn{1}{c|}{RMSE}                & 45.94 $\pm$ 0.57                    & 38.58 $\pm$ 0.70                 & 38.78 $\pm$ 0.58                 & 42.57 $\pm$ 3.31                      & \textit{42.78 $\pm$ 0.07}                & 39.03 $\pm$ 0.27                   & \textbf{35.80$\pm$ 0.18}          \\ \hline
			\multicolumn{1}{c|}{\multirow{3}{*}{\shortstack{PEMS08}}} & \multicolumn{1}{c|}{MAE}                & 22.20 $\pm$ 0.18                   & 17.86 $\pm$ 0.03                  & 18.02 $\pm$ 0.14                  & 18.61 $\pm$ 0.40                      & \textit{19.13 $\pm$ 0.08}                & 17.13 $\pm$ 0.09                   & \textbf{16.64$\pm$ 0.09}          \\  
			\multicolumn{1}{c|}{}                        & \multicolumn{1}{c|}{MAPE(\%)}               & 14.20 $\pm$ 0.59                    & 11.45 $\pm$ 0.03                 & 11.40 $\pm$ 0.10                  & 13.08 $\pm$ 1.00                      & \textit{12.68 $\pm$ 0.57}                & 10.96 $\pm$ 0.07                   & \textbf{10.60$\pm$ 0.06}          \\ 
			\multicolumn{1}{c|}{}                        & \multicolumn{1}{c|}{RMSE}               & 34.06 $\pm$ 0.32                   & 27.83 $\pm$ 0.05                  & 27.83 $\pm$ 0.20                  & 28.16 $\pm$ 0.48                     & \textit{31.05 $\pm$ 0.07}                & 26.80 $\pm$ 0.18                   & \textbf{26.22$\pm$ 0.15}          \\ \hline
	\end{tabular}}
	\caption{Performance comparison of STFGNN and baseline models on PEMS03, PEMS04, PEMS07 and PEMS08 datasets.}
	\label{tab:pems03-08}
\end{table*}

\subsection{Baseline Methods}
We compare STFGNN with those following models:
\begin{itemize}

\item FC-LSTM: Long Short-Term Memory Network, which is a recurrent neural network with fully
connected LSTM hidden units\cite{sutskever2014sequence}.

\item DCRNN: Diffusion Convolution Recurrent Neural Network, which integrates graph convolution into a encoder-decoder gated recurrent unit\cite{li2017diffusion}.

\item STGCN: spatio-temporal Graph Convolutional
Networks, , which integrates graph convolution into a 1D convolution unit\cite{yu2017spatio}.

\item ASTGCN(r): Attention Based Spatial Temporal Graph Convolutional Networks, which introduces spatial and temporal attention mechanisms into model. Only recent components of modeling periodicity is taken to keep fair comparison\cite{guo2019attention}.

\item GraphWaveNet: Graph WaveNet is a framework combines adaptive adjacency matrix into graph convolution with 1D dilated convolution\cite{wu2019graph}.

\item STSGCN: Spatial-Temporal Synchronous Graph Convolutional Networks, which utilizes localized spatial-temporal subgraph module to model localized correlations independently\cite{song2020spatial}.

\end{itemize}

\subsection{Experiment Settings}
To make fair comparison with previous baselines, we split the data with ratio 6 : 2 : 2 at PEMS03, PEMS04, PEMS07, PEMS08 into training sets, validation sets and test sets. One hour 12 continuous time steps historical data is used to predict next hour's 12 continuous time steps data. STFGNN is evaluated more than 10 times in each public dataset.

Experiments are conducted under the environment with one Intel(R) Xeon(R) Gold 6240 CPU @ 2.60GHz and NVIDIA TESLA V100 GPU 16GB card. The temporal graph $A_{TG}$ generated by fast-DTW in Alg. \ref{alg:tgg} costs less than 30 minutes in most public datasets. The Searching Length "T" in "fast-DTW" algorithm is 12, which is the largest prediction time steps in our traffic forecasting task. The sparsity of $A_{TG}$ is 0.01. The model contains 3 STFGNLs, where each contains 8 independent STFGNMs and 1 gated convolution module with dilation rate 3 because the size K of spatial-temporal fusion graph we use is 4. Elements of all three kinds graph are booled to 0 or 1 for the sake of simplification. Filters in each convolution are all 64. We train our model using Adam optimizer with learning rate 0.001. The threshold parameter of loss function $\delta$ is 1, the batch size is 32 and the training epoch is 200. 

\subsection{Experiment Results and Analysis}

Table \ref{tab:pems03-08} shows the through comparison between different models. Results show our STFGNN outperforms baseline models consistently and overwhelmingly on every dataset.

Followed by metrics previous baseline\cite{song2020spatial} takes, Table \ref{tab:pems03-08} compares the performance of STFGNN and other models for 60 minutes ahead prediction on PEMS03, PEMS04, PEMS07 and PEMS08 datasets.

All these four datasets are not particularly smooth, the relatively poor performance of GraphWaveNet reveals its struggle because it can not stack its spatial-temporal layers and enlarge receptive fields of 1D CNN concurrently. 

Modules of STSGCN only extract local spatial-temporal dependencies, and their modules only use multiplication operation (Fully-connected network and adjacency matrix multiplication operation). Thus frequent missing values would disturb its local learning module and smooth time series would magnify its limited representation ability. 

\subsection{Ablation Experiments}

To verify effectiveness of different parts in STFGNN, we conduct ablation experiments on PEMS04 and PEMS08. Table \ref{tab:ablation} shows metric of MAE, MAPE and RMSE. The "Model Element" represents each configuration. Some conclusions could be drawn: 

\begin{itemize}
\item For ingredient of $A_{STFG}$, larger $A_{STFG}$ means more complicated heterogeneity in spatial-temporal dependencies could be extracted regradless of less stacking layers. 

\item For sparsity of $A_{TG}$, it is an important hyperparameter, which determines performance of STFGNN. Empirically, it was set based on sparsity of prior spatial graph. We also demonstrate, with proper sparsity of $A_{TG}$, spatial information free traffic forecasting model is possible, which has promising application value if $A_{SG}$ is unavailable.

\item For Gated Convolution Module, it could remedy long-range learning ability of STFGN Modules which could improve performance of STSGNN.
\end{itemize}

\begin{table}[]
	\begin{tabular}{clccc}
		\hline
		Dataset                 & Model Elements    & MAE            & MAPE\%         & RMSE           \\ \hline
		\multirow{6}{*}{PEMS04}
		& STSGCN        & \underline{21.19}          & \underline{13.90}          & \underline{33.65}          \\
		& $[{ST}_3, T_{sp5}]$     & 20.74          & 13.77          & 33.44          \\
		& $[{ST}_3, T_{sp1}]$     & 20.09          & 13.24          & 32.44          \\
		& $[{ST}_4, T_{sp1}]$           & 19.92          & 13.03          & 31.93          \\
		& $[{T}_4, T_{sp1}, \Theta]$ & 20.02               & 13.17               & 31.98               \\
		& $[{T}_4, T_{sp5}, \Theta]$ & 19.91        & 13.11               & 32.19               \\
		& $[{ST}_4, T_{sp1}, \Theta]$ & \textbf{19.83} & \textbf{13.02} & \textbf{31.88} \\ \hline
		\multirow{6}{*}{PEMS08} & STSGCN        & 17.13          & 10.96          & 26.80          \\
		& $[{ST}_3, T_{sp5}]$      & \underline{19.47}          & \underline{12.27}          & \underline{29.59}          \\
		& $[{ST}_3, T_{sp1}]$     & 16.84          & 10.80          & 26.58          \\
		& $[{ST}_4, T_{sp1}]$          & 16.70          & 10.63          & 26.24          \\
		& $[{T}_4, T_{sp1}, \Theta]$ & 18.23              & 11.52               & 29.05               \\
		& $[{T}_4, T_{sp5}, \Theta]$  & \textbf{16.02}               & \textbf{10.07}               & \textbf{25.39}               \\ 
		& $[{ST}_4, T_{sp1}, \Theta]$  & 16.64 & 10.60 & 26.22 \\ \hline
	\end{tabular}
	\caption{Ablation experiments on different configurations of modules. $ST_4$ means $A_{STFG}$ with size k = 4. $T_4$ means $A_{SG}$ is all replaced to $A_{TG}$ in $A_{STFG}$. $T_{sp5}, T_{sp1}$ means nonzero ratio of $A_{TG}$ is about 5\% and 1\%, respectively. $\Theta$ represents whether gated convolution module is added into each STFGN layer. The default STFGNN configuration we use in this paper is $[ST_4, T_{sp1}, \Theta]$.}
	\label{tab:ablation}
\end{table}

\section{Conclusion}
In this paper, we present a novel framework for spatial-temporal traffic data forecasting. Our model could capture hidden spatial-dependencies effectively by a novel data-driven graph and its further fusion with given spatial graph. By integration with STFGN module and a novel Gated CNN module which enlarges receptive filed on temporal sequences and stacking it, STFGNN could learn localized spatial-temporal heterogeneity and global spatial-temporal homogeneity simultaneously. Detailed experiments and analysis reveal advantages and defects of previous models, which in turn demonstrate STFGNN consistent great performance.

\section{Acknowledgements}
This project is supported by The National Defense Basic Scientific Research Project, China (No. JCKY2018204C004), National Natural Science Foundation  of  China  (No.61806009 and 61932001),  Beijing Nova Program (No. 202072) from Beijing Municipal Science \& Technology Commission and PKU-Baidu Funding 2019BD005.
\bibliography{STFGNN}
\end{document}